\documentclass[runningheads]{llncs}

\usepackage{eccv}

\usepackage{eccvabbrv}

\usepackage{graphicx}
\usepackage{booktabs}
\usepackage{multirow}
\usepackage{paralist}
\usepackage{algpseudocode}
\usepackage{algorithm}
\usepackage[accsupp]{axessibility}  

\usepackage[pagebackref,breaklinks,colorlinks,citecolor=eccvblue]{hyperref}

\usepackage{orcidlink}

\begin{document}

\title{Robust Zero-Shot Crowd Counting and Localization With Adaptive Resolution SAM} 


\titlerunning{Zero-Shot Crowd Counting and Localization}

\author{Jia Wan\inst{1}\orcidlink{0000-0001-8198-1629} \and
Qiangqiang Wu\inst{2}\orcidlink{0000-0002-3847-7838} \and
Wei Lin\inst{2}\orcidlink{0000-0001-8425-956X} \and
Antoni Chan\inst{2}\orcidlink{0000-0002-2886-2513}} 

\authorrunning{J.~Wan et al.}

\institute{School of Computer Science and Technology, Harbin Institute of Technology, Shenzhen \and
Department of Computer Science, City University of Hong Kong \\
\email{jiawan1998@gmail.com, qiangqwu2-c@my.cityu.edu.hk, elonlin24@gmail.com, abchan@cityu.edu.hk}}

\maketitle

\begin{abstract}
The existing crowd counting models require extensive training data, which is time-consuming to annotate. To tackle this issue, we propose a simple yet effective crowd counting method
by utilizing the Segment-Everything-Everywhere Model (SEEM), an adaptation of the Segmentation Anything Model (SAM), to generate pseudo-labels for training crowd counting models.
However, our initial investigation reveals that SEEM's performance in dense crowd scenes is limited, primarily due to the omission of many persons in high-density areas.
To overcome this limitation, we propose an adaptive resolution SEEM to handle the scale variations, occlusions, and overlapping of people within crowd scenes. Alongside this, we introduce a robust localization method, based on Gaussian Mixture Models, for predicting the head positions in the predicted people masks. 
Given the mask and point pseudo-labels, we 
propose a robust loss function, which is designed to exclude uncertain regions based on SEEM's predictions, thereby enhancing the training process of the counting network.
Finally, we propose an iterative method for generating pseudo-labels. 
This method aims at improving the quality of the segmentation masks by identifying more tiny persons in high-density regions, which are often missed in the first pseudo-labeling iteration.
Overall, our proposed method achieves the best unsupervised performance in crowd counting, while also being comparable to some classic supervised fully methods. This makes it a highly effective and versatile tool for crowd counting, especially in situations where labeled data is not available.
  \keywords{Crowd Counting \and Crowd Localization \and Segment Anything}
\end{abstract}
\section{Introduction}
\label{sec:intro}

\begin{figure}[t!]
    \centering
    \includegraphics[width=0.6\textwidth]{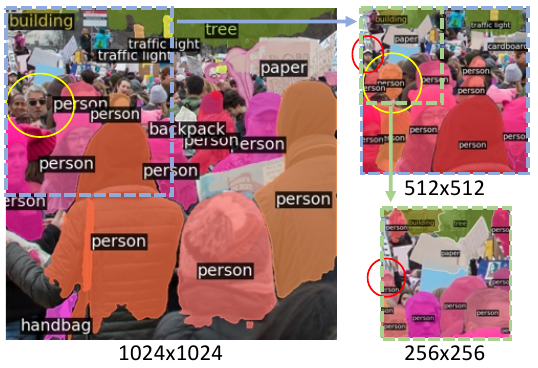}
    \caption{The motivation for our proposed method lies in accurately detecting individuals in high-density areas, where they are often missed due to occlusion and overlapping. Our approach includes zooming into these crowded regions, as this increased resolution helps in identifying previously undetected individuals. For consistency, all regions are resized to $512\times512$ pixels before segmentation. 
    }
    \label{fig:teaser}
\end{figure}

Crowd counting plays a vital role in various applications, from urban planning and public safety to event management and retail \cite{chan2008privacy}. It helps in designing efficient public spaces, optimizing crowd control at events, and managing customer flow in stores. Additionally, it aids in creating responsive infrastructures that adapt to changing population densities. This technology is essential for understanding and managing crowd dynamics in different contexts.

The state-of-the-art crowd counting systems, utilizing deep learning methods like Convolutional Neural Networks (CNNs) \cite{ma2019bayesian} and Transformers \cite{lin2022boosting}, achieve remarkable performance. However, these methods typically require substantial amounts of labeled data for training. The scale of crowd counting datasets is relatively small, as labeling each person in dense crowd images is a time-consuming task.
As a result, there is a growing need for unsupervised methods capable of adapting to new datasets without relying on manual annotations.

To tackle this challenge, we introduce a robust unsupervised method that utilizes the Segmentation Anything Model (SAM) \cite{Kirillov_2023_ICCV} to generate pseudo labels. 
However, SAM is not able to predict semantic labels. Therefore, the Segment-Everything-Everywhere Model (SEEM)~\cite{zou2023segment} is utilized to predict person masks. 
Large foundation models are shown to be useful for downstream tasks \cite{zhu2023trackinghumanintentreasoning}. However, our findings indicate that using SEEM directly is not effective, since it often misses people due to occlusions and overlapping (see 1024x1024 image in Fig.~\ref{fig:teaser}), which is due to the limited availability of dense crowd images in its training data. To address this, we propose an adaptive resolution SEEM (AdaSEEM) that can zoom in on areas of high density as needed. This enhancement allows for more precise segmentation of smaller persons in crowded regions as shown in Figure~\ref{fig:teaser}. In addition, we propose a robust head localization method to estimate the head locations accurately by modeling the mask distribution as a Gaussian Mixture Model (GMM), enabling the generation of more effective point pseudo-labels. 

We use the generated mask and point pseudo-labels to train a counting regression network.
To effectively use both types of pseudo-labels, we propose a robust loss function composed of two parts: an individual loss and a background loss, with uncertain regions excluded during training. The individual loss ensures that the total density within a mask is close to 1, and it also encourages the density to converge around the head pseudo-labels. This approach enhances the accuracy of crowd counting, as well as ensures precise localization within segmented areas. The background loss, in contrast, is tailored to predict a zero value for all background regions, thereby efficiently reducing false positive predictions in non-crowded areas. 

Finally, to enhance performance, we adopt an iterative approach for generating pseudo masks, using the point predictions from the well-trained counting network as prompts for AdaSEEM. This helps in identifying missing individuals in high-density areas. Once these new masks are created, they are fused with those from the previous iteration to create a more comprehensive and accurate set of pseudo-labels. Subsequently, we employ the same methodology to estimate head point pseudo-labels within these updated masks. With these refined masks and head locations, we proceed to train the counting networks, thereby improving their accuracy and reliability in densely populated scenes.

In summary, the paper has four key contributions: 
\begin{compactenum}
\item We introduce a novel approach for generating both mask and point pseudo labels for unsupervised crowd counting. This involves the use of the Segmentation Anything Model (SAM) enhanced with an adaptive resolution strategy and a robust mechanism for localizing head points.
\item To leverage both mask and point pseudo labels, we develop a robust loss function that strategically excludes uncertain regions during training, and ensures the density within each mask is 1. This function is instrumental in accurately counting and localizing individuals within crowded scenes.
\item We propose an iterative method for pseudo mask generation. This approach refines mask predictions by utilizing point prompts derived from the currently-trained counting network, allowing for the identification of previously missed individuals in dense areas.
\item Our method significantly outperforms existing unsupervised crowd counting methods, showing improvements by a large margin. Its performance is also comparable to some classic fully supervised methods, even on large-scale datasets.
\end{compactenum}

\section{Related Works}
\label{sec:related}

In this section, we briefly review the supervised, semi-supervised and unsupervised crowd counting algorithms.

\subsection{Supervised Methods}

Traditional crowd counting algorithms rely on individual detection \cite{ge2009marked}, which does not generalize well to high-density images due to occlusion. To improve counting performance, direct regression methods have been proposed that utilize low-level features \cite{chan2008privacy}, including texture \cite{chan2009bayesian} and color \cite{idrees2013multi}. However, the effectiveness of these methods is still limited by factors like scale and scene variation.

Recent crowd counting research has predominantly focused on deep learning, with significant improvements achieved through training with extensive labeled data \cite{zhang2016single,huang2020stacked,zhang2022crossnet,cheng2022rethinking,wang2024ivacp2lleveragingirregularrepetition}. Innovations in network structures \cite{babu2017switching,han2023steerer,cheng2019improving} and the development of various loss functions \cite{wang2020distribution} have enhanced performance and robustness.  \cite{kang2018crowd} introduced the use of an image pyramid to address scale variation. Further enhancements include the exploitation of contextual information \cite{sindagi2017generating,xiong2017spatiotemporal,cheng2019learning} and the development of cross-scene crowd counting methods to improve generalization \cite{zhang2015cross}. \cite{wang2019learning} proposed the use of a synthetic dataset, while others have explored the use of correlation information to boost generalization capabilities \cite{wan2019residual,wu2021dynamic}. Innovative approaches in loss function design, such as learnable density maps for enhanced supervision, have been proposed \cite{wan2019adaptive,wan2020kernel}. Direct use of point annotations during training has shown improved counting and localization \cite{wang2020distribution,song2021rethinking,ma2021learning,liu2023point}, and robust loss functions have been developed to address annotation noise \cite{wan2020modeling,10197253}. Recently, Transformer-based methods have demonstrated exceptional performance in both crowd counting and localization \cite{lin2022boosting,liang2022end}.

However, supervised methods require a significant quantity of labeled images, which can be challenging to acquire due to the time-consuming labeling process, e.g., some training images may contain hundreds or even thousands of people. In contrast, our proposed unsupervised method attains results comparable to those achieved by some supervised methods, without requiring any labeled crowd images. 

\subsection{Semi-Supervised And Unsupervised Methods}
To alleviate the burden of extensive annotation, several innovative approaches have been proposed in the realm of crowd counting \cite{10251149}. \cite{change2013semi} suggest the use of unlabeled videos, thereby reducing the dependency on fully labeled datasets.  \cite{meng2021spatial} introduce a method to model spatial uncertainty, enhancing the efficacy of semi-supervised counting. The concept of training models with partial annotations has also been explored \cite{xu2021crowd}, offering a practical alternative to fully supervised methods. Furthermore, a supervised uncertainty estimation strategy is presented in \cite{LI_2023_ICCV}, providing a novel approach to address annotation challenges. Additionally, the use of optimal transport minimization \cite{lin2023optimal} has been proposed for crowd localization in semi-supervised settings, further contributing to the development of more efficient and less labor-intensive methods in the field of crowd counting.

The exploration of unsupervised crowd counting methods, especially for high-density scenarios, remains limited. Most existing research in this area tends to concentrate on low-density images. 
A novel self-supervised method based on distribution matching has been proposed in \cite{babu2022completely}. Additionally, \cite{liang2023crowdclip} introduced an innovative approach by employing Vision-Language models for zero-shot crowd counting. 
%
%
While these unsupervised methods demonstrate reasonably good performance, their effectiveness in high-density scenes is still not optimal. In contrast, our proposed method stands out by achieving performance levels comparable to some supervised methods, even in complex, high-density environments, thus 
offering a viable alternative to traditional supervised approaches that require extensive labeled data.

\begin{figure*}[h]
    \centering
    \includegraphics[width=0.95\textwidth]{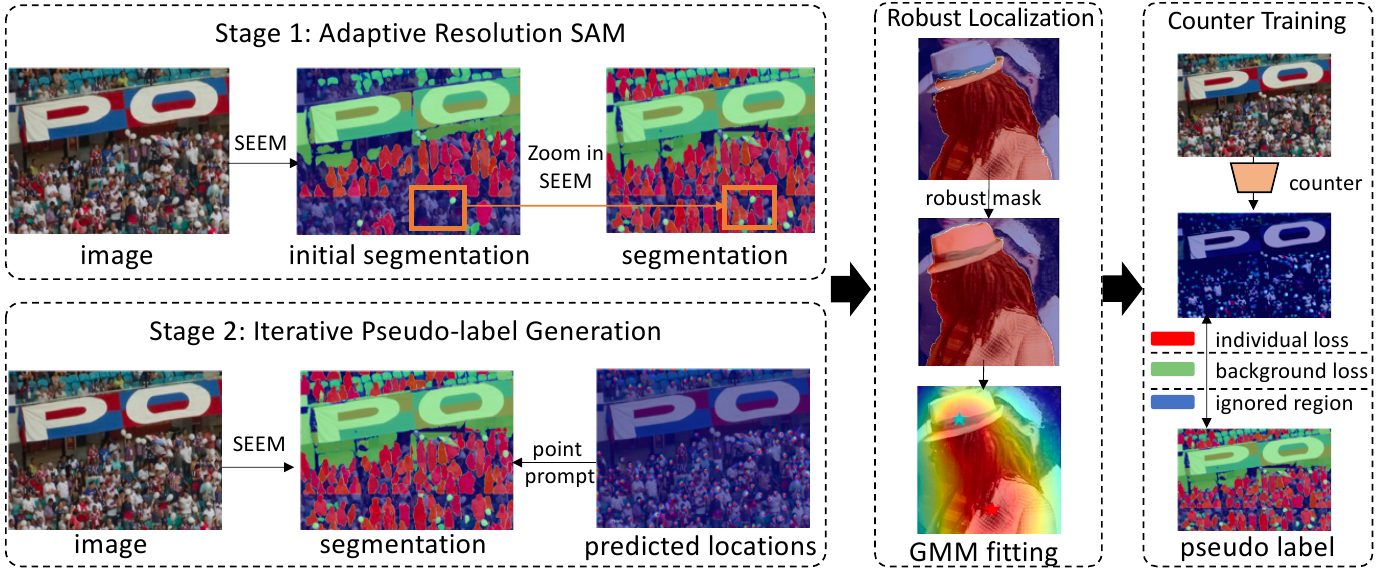}
    \caption{Our framework for unsupervised crowd counting. 
    First, we generate person mask pseudo-labels using an adaptive resolution SAM (AdaSEEM) to enhance the segmentation of small-sized objects in crowd images. We then predict point pseudo-labels via a robust method for head localization achieved by modeling the soft mask distribution using a Gaussian Mixture Model (GMM). The next phase involves training a counting network using a robust loss function that is specifically designed to use the generated mask/point pseudo labels.  Finally, we employ an iterative process to generate additional pseudo labels by leveraging the predictions of the trained counting network.}
    \label{fig:pipeline}
\end{figure*}

\section{Method}
\label{sec:method}

In this paper, we introduce a novel robust unsupervised crowd counting method that harnesses the capabilities of the Segmentation Anything Model (SAM). Our approach consists of several key steps. 
We first propose an adaptive inference strategy for utilizing SAM, which 
enables more precise segmentation of individuals, especially those of smaller sizes, in various crowd scenes.
We then introduce a robust method for localizing head positions within the predicted individual masks. This step is crucial for obtaining precise point pseudo-annotations for more accurate counting.
Utilizing the  masks and point pseudo-labels generated, we train a counting network. Our training process is distinguished by a robust loss function that deliberately excludes uncertain regions, thereby enhancing the model's precision and reliability.
Finally, we propose an iterative process for generating pseudo-labels. This process is based on the predictions of the counting network, and aims at continuously improving the quality of the pseudo labels.
The overall workflow of our proposed method is  illustrated in Figure~\ref{fig:pipeline}.

\subsection{Adaptive Resolution SAM}
SAM, initially designed for generic segmentation tasks, has been trained on millions of images, which grants it an impressive ability to generalize across various scenarios. However, a key limitation of SAM is its inability to assign specific object categories to the segments it identifies. To overcome this, we opt for a modified version of SAM, known as the Segment-Everything-Everywhere Model (SEEM) \cite{zou2023segment}. SEEM, having been trained with semantic labels, is adept at providing a semantic label for each mask, enhancing its utility in segmentation tasks.
Despite its capabilities, SEEM faces challenges in detecting small individuals in crowded images. This limitation primarily arises due to the relatively small proportion of dense crowd images in its training dataset \cite{Kirillov_2023_ICCV}. To address this specific issue, we introduce an adaptive resolution SEEM (denoted as AdaSEEM). This strategy is designed to improve the model's performance in identifying small-sized persons in high-density crowd scenes, thus enhancing the overall effectiveness and applicability of SEEM for generating mask pseudo-labels in complex crowd counting scenarios.

In our approach, we initially apply SEEM to the original image to obtain segmentation results. These results are categorized into three distinct groups: non-person (background) regions, uncertain regions, and individual person masks as shown in Figure~\ref{fig:pipeline}.
The non-person background regions are the segments with non-person labels, while the uncertain region contains pixels that do not belong to any segment. 
Following the initial segmentation, we crop the image into smaller patches and assess the proportion of uncertain regions in each patch. If a patch has an uncertain region ratio exceeding a predefined threshold $\tau$, we then zoom into this patch, doubling its resolution, and reapply SEEM. 
The Non-Maximum Suppression (NMS) is used to merge segments from different iterations.
This process is iterative and continues until the ratio of uncertain regions in all patches falls below the threshold. 
By iteratively zooming in and reapplying SEEM on patches with high uncertainty, we significantly improve the accuracy of our segmentation, especially in detecting smaller individuals in dense crowd scenes. This adaptive approach ensures that the segmentation results are both precise and reliable, increasing their effectiveness as pseudo-labels for crowd-counting. 

\subsection{Robust Localization for Point Pseudo-labels}\label{subsec:loc}
Crowd counting methods typically require point annotations for training. Thus, we propose an algorithm to predict the head location from each individual person mask generated by AdaSEEM.
Our approach begins with the generation of a robust mask distribution from the initial mask. Denote the predicted initial mask as $M_0$. We randomly sample $K$ points in $M_0$ and use these as prompts to SEEM to generate new masks, denoted as $\{M_n\}_{n=1}^K$.
We then compute the soft mask distribution, by averaging over the predicted masks: $M = \frac{1}{K+1}\sum_{i=0}^{K} M_i$. This averaging process helps in smoothing out the noise and inconsistencies in the initial mask predictions.

Inspired by classic density map generation \cite{zhang2016single}, we then model the soft mask distribution $M$ using a Gaussian Mixture Model (GMM) with two components. The model is represented as follows:
\begin{equation}\label{eq:gmm}
p(x) = \sum_{i=1}^{2} \pi_i \mathcal{N}(x | \mu_i, \Sigma_i),
\end{equation}
where $\mu_i$ and $\Sigma_i$ represent the mean and variance of each Gaussian distribution within the mixture.

We fit the soft mask distribution $M$ to the GMM with the expectation-maximization (EM) algorithm (see Supplemental).
The final step involves selecting the mean $\mu_i$ of the Gaussian component with the smaller vertical coordinate (height) as the head location. This method effectively utilizes the statistical properties of the GMM to pinpoint the head location, thus accommodating the variability and noise in the segmentation process.

\subsection{Counter Training with Robust Loss}\label{subsec:train_counter}
The counting network is trained using the generated mask and point pseudo-labels. For an input image $I$, the corresponding pseudo label consists of the background mask $ M_b $, the uncertain mask $ M_u $, individual masks $ \{M_i\}_{i=1}^N$, and head locations $\{p_i\}_{i=1}^N$, where $ N $ is the number of annotated people in the image.

Our proposed loss function for the predicted density map $ \hat{D} $ comprises two components: a background loss and an individual loss, with predictions in uncertain regions being disregarded.
The background loss is defined for the background (non-person) regions, where the prediction should be close to 0. It is formulated as follows:
\begin{equation}
\mathcal{L}_{bkg} = \langle \hat{D}, ~M_b \rangle, \label{eq:lb} 
\end{equation}
where $\langle \cdot , \cdot \rangle$ means performing component-wise inner product of two vectorized matrices.

The individual loss is given by:
\begin{equation}
\mathcal{L}_{idv} = \frac{1}{N}\sum_{i=1}^N \Big[ \big| \langle \hat{D}, ~M_i \rangle - 1 \big| + \omega \big\langle \tfrac{\hat{D} ~\circ~ M_i}{\langle \hat{D}, ~M_i\rangle}, ~C_i\big\rangle\Big], \label{eq:li}
\end{equation}
where $C_i$ is an exponential distance matrix, in which the $j$-th element $C_{i}^{[j]} = \exp(-\|x_j-p_i\|^2/\epsilon) $ represents the exponential distance between the head location $p_i$ and the density value location $x_j$. $\circ$ is the element-wise product. The second term encourages the density to converge towards the head. For more details on this, please refer to \cite{wan2021generalized}.

The final loss function is a combination of the background loss in (\ref{eq:lb}) and individual loss in (\ref{eq:li}):
\begin{equation}
\mathcal{L} = \mathcal{L}_{idv} + \beta \mathcal{L}_{bkg}, 
\end{equation}
where $\beta$ is a weighting hyperparameter. 

\begin{table*}[!h]
\centering
\resizebox{\textwidth}{!}{
\begin{tabular}{lcccccccccccc}
\hline
\multirow{2}{*}{\textbf{Method}} & \multirow{2}{*}{\textbf{Year}} & \multirow{2}{*}{\textbf{Label}} & \multicolumn{2}{c}{\textbf{UCF-QNRF}} & \multicolumn{2}{c}{\textbf{JHU}} & \multicolumn{2}{c}{\textbf{ShTech A}} & \multicolumn{2}{c}{\textbf{ShTech B}} & \multicolumn{2}{c}{\textbf{UCF-CC-50}} \\
& & & \textbf{MAE} & \textbf{MSE} & \textbf{MAE} & \textbf{MSE} & \textbf{MAE} & \textbf{MSE} & \textbf{MAE} & \textbf{MSE}& \textbf{MAE} & \textbf{MSE}  \\ 
\hline
Zhang et al. \cite{zhang2015cross} & CVPR 15 & Point & - & - & - & - & 181.8 & 277.7 & 32.0 & 49.8&467.0 &498.5 \\
MCNN \cite{zhang2016single} & CVPR 16 & Point & 277.0 & 426.0 & 188.9 & 483.4 & 110.2 & 173.2 & 26.4 & 41.3 & 377.6 & 509.1\\
Switch CNN \cite{babu2017switching} & CVPR 17 & Point & 228.0 & 445.0 & - & - & 90.4 & 135.0 & 21.6 & 33.4  &318.1& 439.2\\
LSC-CNN \cite{sam2020locate} & TPAMI 21 & Point & 120.5 & 218.2 & 112.7 & 454.4 & 66.4 & 117.0 & 8.1 & 12.7  &225.6 &302.7\\
SDA+DM \cite{ma2021towards} & ICCV 21 & Point & 80.7 & 146.3 & 59.3 & 248.9 & 55.0 & 92.7 & - & - &-&-\\ 
CLTR \cite{liang2022end} & ECCV 22 & Point & 85.8 & 141.3 & 59.5 & 240.6 & 56.9 & 95.2 & 6.5 & 10.2  &-&-\\
MAN \cite{lin2022boosting} & CVPR 22 & Point & 77.3 & 131.5 & 53.4 & 209.9 & 56.8 & 90.3 & - & - &-&-\\
Chfl \cite{shu2022crowd} & CVPR 23 &Point & 80.3 & 137.6 & 57.0 & 235.7 & 57.5 & 94.3 & 6.9 & 11.0&-&- \\
STEERER \cite{han2023steerer} & ICCV 23 & Point &74.3&128.3&54.3&238.3&54.5&86.9&5.8&8.5 &-&-\\
\hline
 SFCN~\cite{wang2019learning} & CVPR 19 & Point (GCC~\cite{wang2019learning}) & 275.5&458.5&-&-&160.0&216.5&22.8&30.6 &487.2& 689.0\\
RCC~\cite{hobley2022learning}	&arXiv 22 & Point (FSC~\cite{ranjan2021learning}) &-&-&-&-&240.1&366.9&66.6&104.8 &-&-\\
CLIP-Count~\cite{jiang2023clip}	& arXiv 23 & Point (FSC~\cite{ranjan2021learning}) &-&-&-&-&192.6&308.4&45.7&77.4 &-&-\\
\hline
CSS-CNN-Rnd. \cite{babu2022completely} & ECCV 22 & None & 718.7 & 1036.3 & 320.3 & 793.5 & 431.1 & 559.0 & - & - & 1279.3 &1567.9 \\
Random* & - & None & 633.6 & 978.9 & 297.5 & 801.6 & 411.1 & 511.1 & 158.7 & 287.4 &1251.6 &1497.8\\
CSS-CNN \cite{babu2022completely} & ECCV 22 & None & 437.0 & 722.3 & 217.6 & 651.3 & 179.3 & 295.9 & - & - &564.9 &959.4\\
CrowdCLIP \cite{liang2023crowdclip} & CVPR 23 & {None} & {283.3} & {488.7} & {213.7} & {576.1} & {146.1} & {236.3} & {69.3} & {85.8}& 438.3 & 604.7 \\
Ours (Iter. 0) &&None&{195.9}&{343.0}&{109.5}&{428.7}&{125.4}&{226.7}&\underline{34.4}&{55.2} &424.5& 597.1\\
Ours (Iter. 1) &&None&\textbf{181.2}&\underline{304.7}&\underline{105.1}&\underline{390.5}&\underline{122.8}&\underline{217.8}&\textbf{33.3}&\underline{53.2} &\underline{382.7}&\textbf{444.5}\\
Ours (Iter. 2) & & None &\underline{182.3}&\textbf{289.9}&\textbf{102.7}&\textbf{360.7}&\textbf{102.6}&\textbf{176.3}&35.6&\textbf{51.7}&\textbf{376.6}&\underline{578.2} \\
\hline
\end{tabular}
}
\caption{Comparison with state-of-the-art methods. ``Point'' label indicates using point annotations as supervision while ``None'' is the unsupervised setting (no crowd labels are used). ``Point (X)'' indicates the method was trained on dataset X (cross-domain performance).  The best unsupervised method is bolded, and 2nd best is underlined.}
\label{table:sota}
\end{table*}

\subsection{Iterative Pseudo-label Generation}\label{subsec:iterative}
One of the key advantages of our proposed method is its capability to predict both the global count and the precise location of each individual within a crowd via a predicted density map (c.f., \cite{liang2023crowdclip} that only predicts the count). This functionality allows for further refinement of the pseudo-labels, especially in finding missed individuals in high-density regions.

The process begins with predicting the locations of individuals using the pretrained counting network. In particular, the local maxima above a threshold are potential person localizations, following \cite{wan2021generalized}. These predicted locations are then used as point prompts to generate new masks using SEEM. To ensure high recall, we use multiple points to generate more masks and then combine duplicate masks with Non-Maximum Suppression (NMS). 
In the subsequent step, these newly generated masks are combined with the masks from the previous iteration using NMS. This iterative strategy is particularly effective in high-density areas, where it can uncover individuals who may have been missed in earlier iterations.

\begin{figure}[t]
    \includegraphics[width=0.8\textwidth]{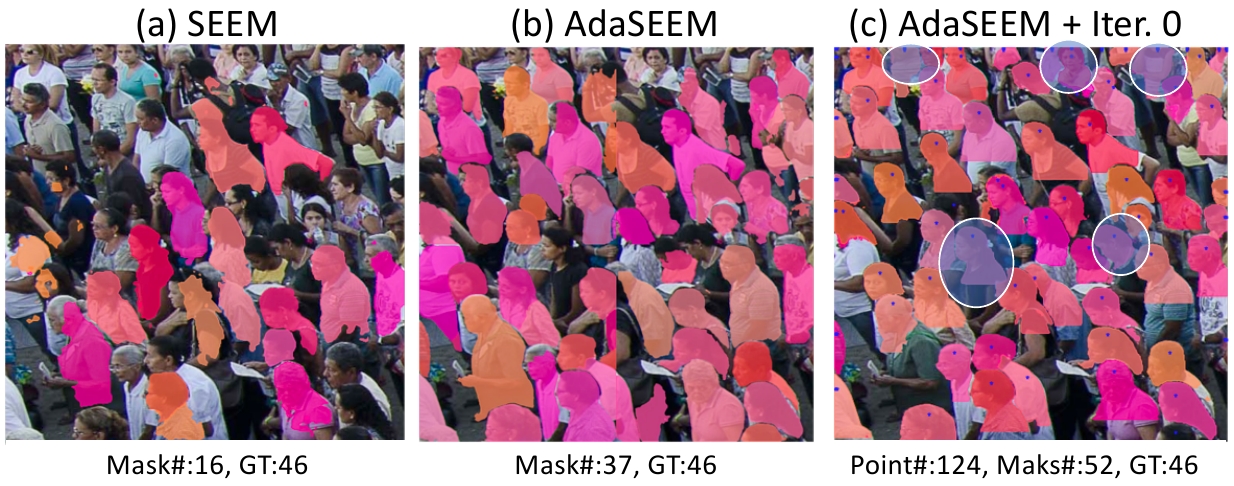}
    \centering
    \caption{The masks generated from different methods. From left to right are: SEEM, adaptive resolution SEEM (AdaSEEM), and AdaSEEM + Iter. 0 predictions. In (c), the new pseudo-label masks are highlighted with blue ellipses.} 
    \label{fig:masks}
\end{figure}

This approach is visually demonstrated in Figure~\ref{fig:masks}, which shows the effectiveness of this strategy in detecting more individuals in densely populated regions. The overall algorithm is summarized in Algorithm~\ref{alg:adaseem}.

\begin{algorithm}[h]
\caption{Unsupervised Crowd Counting with Robust AdaSEEM}\label{alg:adaseem}
\begin{algorithmic}
\Require SEEM model, unlabeled training images $\{I_i\}$ 
\State \emph{\scriptsize \texttt{\# Generate pseudo-masks with AdaSEEM}}
\State $\{M_i\}_i=\{\text{SEEM}(I_i)\}_i$ \Comment{Segment with SEEM} 
\For{each image $I_i$}
\State $s=512$
\While{(uncertain ratio in $M_i$ $>$ $\tau$) and ($s \geq 64$)} 
\State split $I_i$ into $s\times s$ patches $\{I_i^j\}_j$
\State $\{I_i^j\}_j=\{\text{zoomin}(I_i^j)\}_j$ \Comment{Zoom in}
\State $\{M_j\}_j=\{\text{SEEM}(I_i^j)\}_j$ \Comment{Segment with SEEM}
\State $\{M_i\} = \text{merge}(\{M_i\}, \{M_j\})$ \Comment{NMS}
\State $s=s \div 2$
\EndWhile
\EndFor
\State \emph{\scriptsize \texttt{\# Generate pseudo-points}}
\State $\{P_i\}_i=\{\text{robustlocalize}(M_i)\}_i$ \Comment{\S\ref{subsec:loc}}
\State \emph{\scriptsize \texttt{\# Train counter}}
\State $Counter=\text{train}(\{I_i\}, \{M_i\}, \{P_i\})$ \Comment{\S\ref{subsec:train_counter}}
\State \emph{\scriptsize \texttt{\# Iterative refinement}}
\For{$k \in \{1,2\}$ } \Comment{Iteration 1\&2}
\State \emph{\scriptsize \texttt{\# Add new masks using point prompts}}
\For{each image $I_i$}
\State $\{\hat{P}_n\}_n = \text{localize}(Counter(I_i))$ \Comment{\S\ref{subsec:iterative}}
\State $\{M_n\}_n=\{\text{SEEM}(I_i, \hat{P}_n)\}_n$ \Comment{Prompt w/ points}
\State $\{M_i\}=\text{merge}(\{M_i\}, \{M_n\})$ \Comment{NMS}
\EndFor
\State \emph{\scriptsize \texttt{\# Generate new pseudo-points}}
\State $\{P_i\}_i=\{\text{robustlocalize}(M_i)\}_i$ \Comment{\S\ref{subsec:loc}}
\State \emph{\scriptsize \texttt{\# Train Iteration-$k$ counter}}
\State $Counter=\text{train}(\{I_i\}, \{M_i\}, \{P_i\})$ \Comment{\S\ref{subsec:train_counter}}
\EndFor
\State \Return $Counter$
\end{algorithmic}
\end{algorithm}

\begin{figure}[t!]
    \centering
    \includegraphics[width=0.8\textwidth]{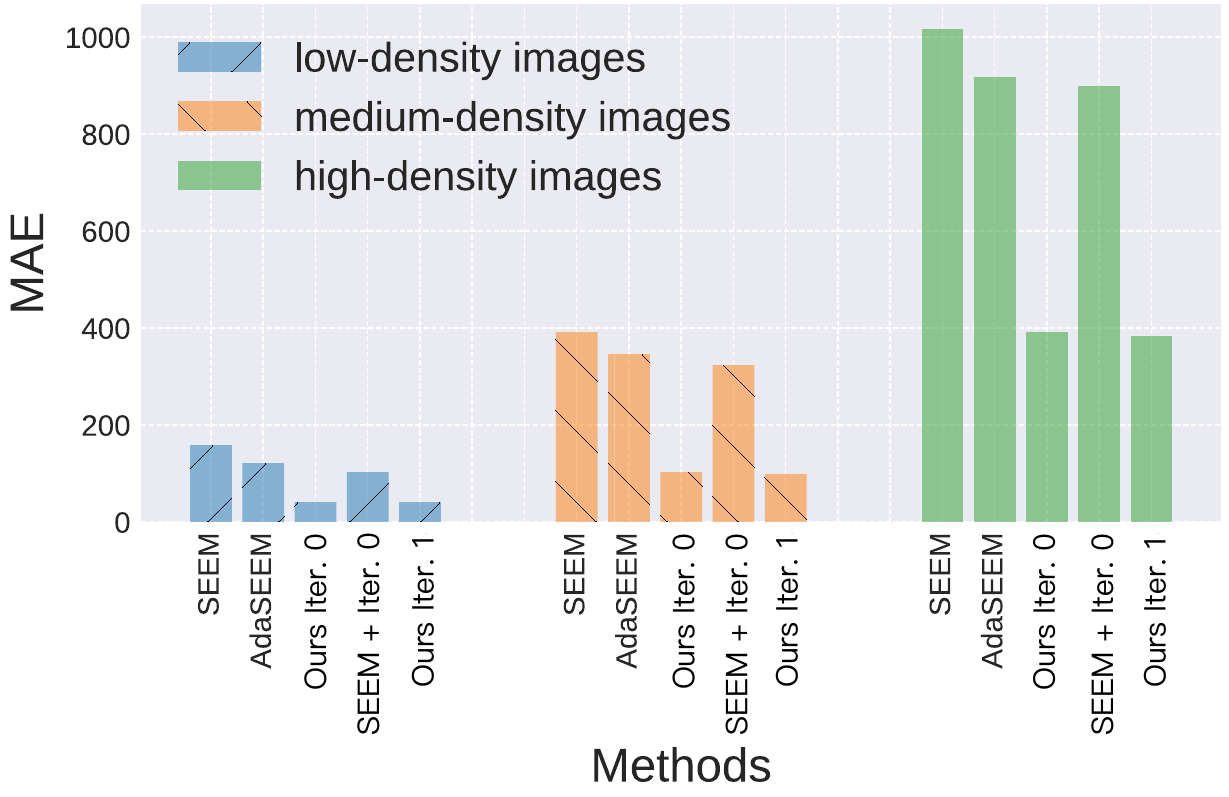}
    \caption{The comparison of different methods across varying density levels of ShanghaiTech A dataset: low-density (count $\leq$ 300), medium-density (300 $<$ count $\geq$ 600), and high-density (count $>$ 600).}
    \label{fig:levels}
\end{figure}

\section{Experiments}
\label{sec:exp}

In this section, we first present the experimental settings. Then, we compare the proposed method with SOTA methods. Finally, different components of the proposed method are evaluated in ablation studies.  

\subsection{Experimental Settings}
\textbf{Dataset:} We evaluate the proposed method on JHU-CROWD dataset \cite{sindagi2020jhu}, UCF-QNRF \cite{idrees2018composition}, ShanghaiTech \cite{zhang2016single} and UCF-CC-50 \cite{idrees2013multi} datasets. 
The JHU-CROWD dataset is a comprehensive large-scale dataset, comprising 4,371 images. It is divided into three subsets: 2,722 images for training, 500 for validation, and 1,600 for testing. The UCF-QNRF dataset includes 1,535 images, with 1,201 designated for training and 334 for testing. The ShanghaiTech dataset is split into two parts: the ShanghaiTech A, which contains a total of 782 images, divided into 482 for training and 300 for testing, and the ShanghaiTech B, which includes 1,116 images, with 716 used for training and 400 for testing. UCF-CC-50 contains 50 grayscale images and we use 5-fold cross-validation in experiments. 

\textbf{Training details:} For our experiments, we employ the counting network architecture from \cite{wan2021generalized}, which is based on the VGG backbone \cite{li2018csrnet}. The network is trained using the Adam optimizer, with a learning rate of 1e-5. We maintain a batch size of 1 across all experiments to ensure consistent training conditions. The models undergo training for a total of 100 epochs, allowing for adequate learning and adaptation to the dataset characteristics. As an unsupervised approach, we do not use the crowd annotations during training, but instead generate pseudo-labels from the training images. 

The parameters $\omega$ and $\beta$ in our loss function play crucial roles in optimizing performance. These parameters are set to 100 and 0.01, respectively, based on the ablation study in Figures~\ref{fig:omega} and \ref{fig:beta}. The threshold $\tau$ in AdaSEEM is set to 0.3 according to the experimental result shown in Figure~\ref{fig:tau}.

\textbf{Metrics:} Following previous works \cite{wan2021generalized}, we use MAE and MSE as the metrics to evaluate the counting performance:
\begin{equation}
    MAE = \frac{1}{N} \sum \|\hat{y_i} - y_i\|,  MSE = \sqrt{\frac{1}{N} \sum \|\hat{y_i} - y_i\|^2},
\end{equation}
where $\hat{y}$ and $y$ are predicted count and the ground-truth count and $N$ is the number of images.

\subsection{Comparison with State-of-the-art Methods}
To assess the effectiveness of our proposed method, we conducted a thorough evaluation by comparing it with both state-of-the-art unsupervised and supervised methods. The results of this comparison are detailed in Table~\ref{table:sota}. First, our proposed method outperforms other unsupervised methods in terms of MAE and MSE, and the margin of improvement is significant. This underscores the effectiveness of our approach in addressing the challenges inherent to unsupervised counting. Second, the comparison also reveals that the performance in Iter. 1 of our method is better than in Iter. 0 across all datasets. This improvement validates the effectiveness of our iterative pseudo-label generation strategy. By refining the pseudo-labels, the model is able to achieve more accurate and reliable counting results.
We also compare the proposed method with cross-domain methods, 
which train on a source dataset and test on the target dataset, 
in Table~\ref{table:sota} and achieve superior performance for most of the cases.
Finally, the proposed method also compares favorably with some classic supervised methods. However, there is still considerable room for improvement, particularly in handling densely crowded datasets.

\begin{figure*}[t]
    \centering
    \includegraphics[width=0.95\textwidth]{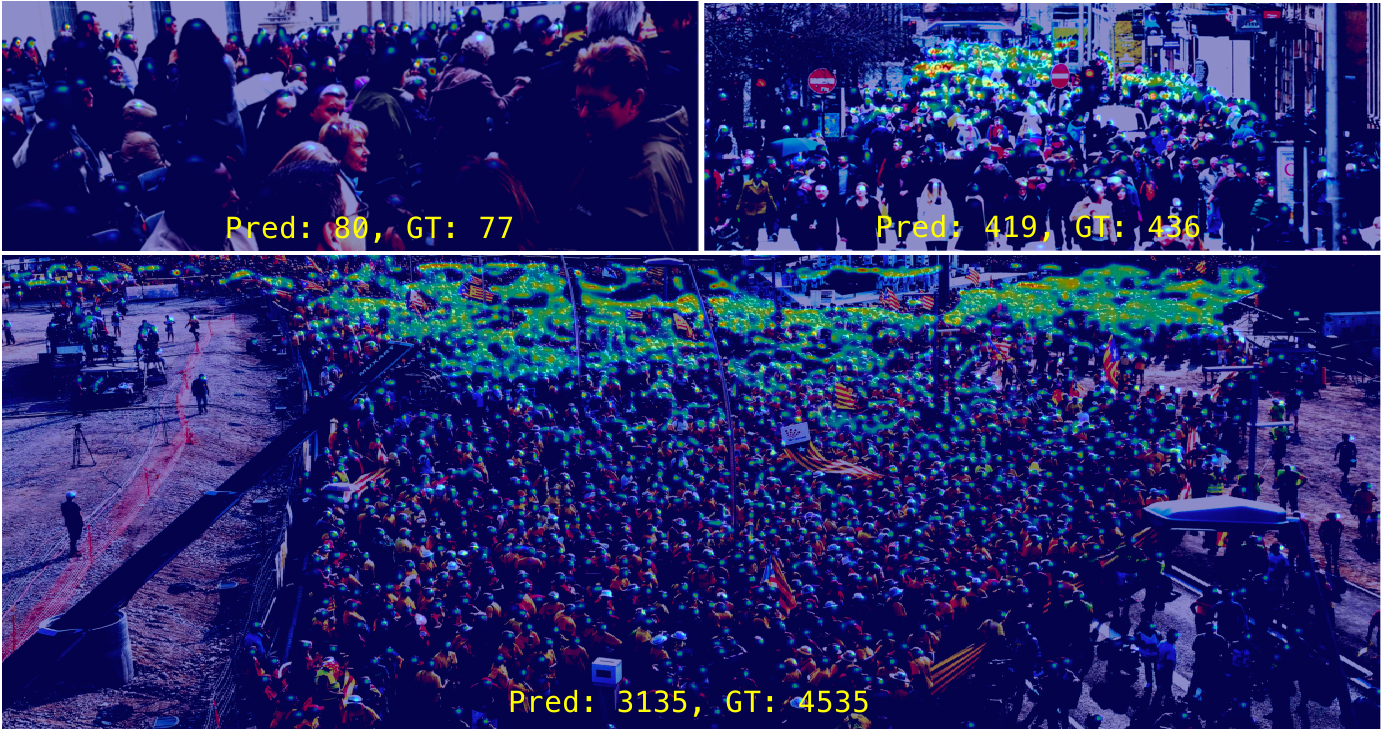}
    \caption{The visualization of the predicted density maps. Note that unsupervised methods typically lack the capability to predict such density maps, e.g., \cite{liang2023crowdclip} only predicts the count.}
    \label{fig:vis}
\end{figure*}

Our unsupervised method effectively predicts density maps from images, as illustrated in Figure~\ref{fig:vis}, enabling precise person location prediction without manual labels. This capability also facilitates the application of our iterative pseudo-labels generation method, enhancing mask quality and overall performance.

\subsection{Ablation Study}

\textbf{Adaptive resolution SAM}
We conducted counting experiments to evaluate the effectiveness of AdaSEEM, and the results are presented in Table~\ref{tab:ablation}. The performance of SEEM on its own was found to be the least effective, indicating that directly using SEEM is not optimal due to the omission of many small individuals in high-density areas. However, with the implementation of the proposed adaptive resolution strategy, there was a noticeable performance improvement, especially for the high-density dataset UCF-QNRF. In Figure~\ref{fig:levels}, we can also observe significant improvement in high-density images when using AdaSEEM. This improvement underscores the efficacy of the adaptive resolution strategy in accurately segmenting small persons in densely populated regions. The effectiveness of this approach is further confirmed in Figure~\ref{fig:masks}(a, b), where more individuals are segmented with the AdaSEEM compared to the base model. These findings collectively highlight the significance of the adaptive resolution strategy in enhancing the segmentation capabilities of SEEM in complex crowd counting scenarios.

\begin{table}[!t]
\centering
\resizebox{0.6\textwidth}{!}{
\begin{tabular}{lcccc}
\hline
\multirow{2}{*}{\textbf{Method}}& \multicolumn{2}{c}{\textbf{ShTech A}} & \multicolumn{2}{c}{\textbf{UCF-QNRF}} \\ 
& MAE & MSE & MAE & MSE \\ \hline 
SEEM & 394.2 & 529.8 &526.4&872.8\\
AdaSEEM &342.9&484.2 &391.2&654.5\\
Ours (Iter. 0) &125.4&226.7 &195.9&343.0\\
AdaSEEM + Iter. 0 &323.4&470.8 &347.5&612.1\\
Ours (Iter. 1) &\textbf{122.8}&\textbf{217.8}&\textbf{181.2}&\textbf{304.7} \\
\hline
\end{tabular}
}
\caption{Ablation studies on ShanghaiTech A and UCF-QNRF datasets.}
\label{tab:ablation}
\end{table}

\begin{figure*}[ht]
    \centering
    \begin{minipage}{0.3\textwidth}
        \centering
        \includegraphics[width=\textwidth]{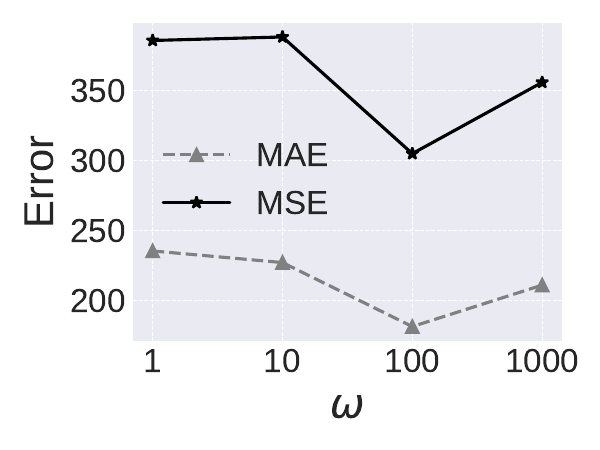} 
        \caption{Error v.s. $\omega$.}
        \label{fig:omega}
    \end{minipage}
    \begin{minipage}{0.3\textwidth}
        \centering
        \includegraphics[width=\textwidth]{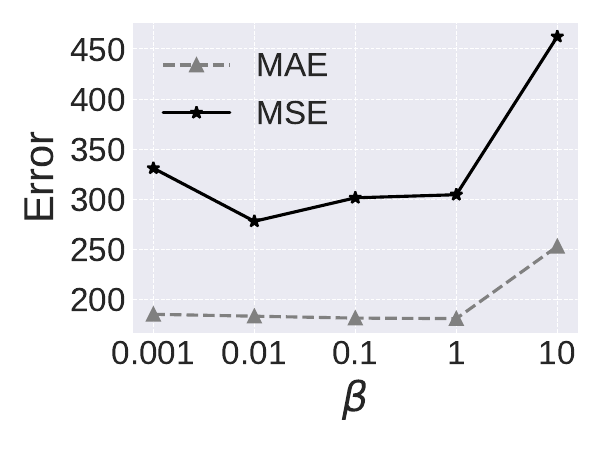} 
        \caption{Error v.s. $\beta$.}
        \label{fig:beta}
    \end{minipage}
    \begin{minipage}{0.3\textwidth}
        \centering
        \includegraphics[width=\textwidth]{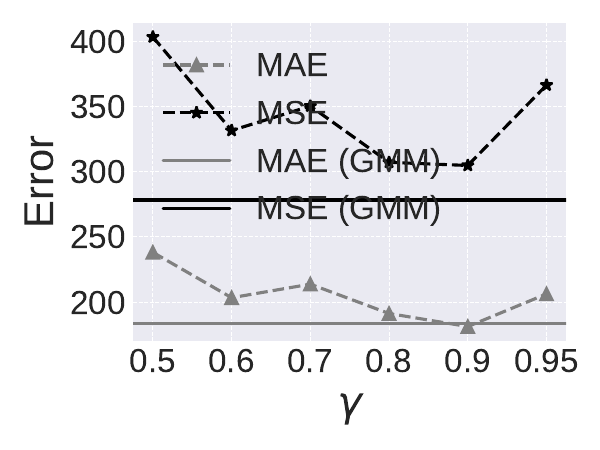} 
        \caption{Error v.s. $\gamma$.}
        \label{fig:gamma}
    \end{minipage}
    \begin{minipage}{0.3\textwidth}
        \centering
        \includegraphics[width=\textwidth]{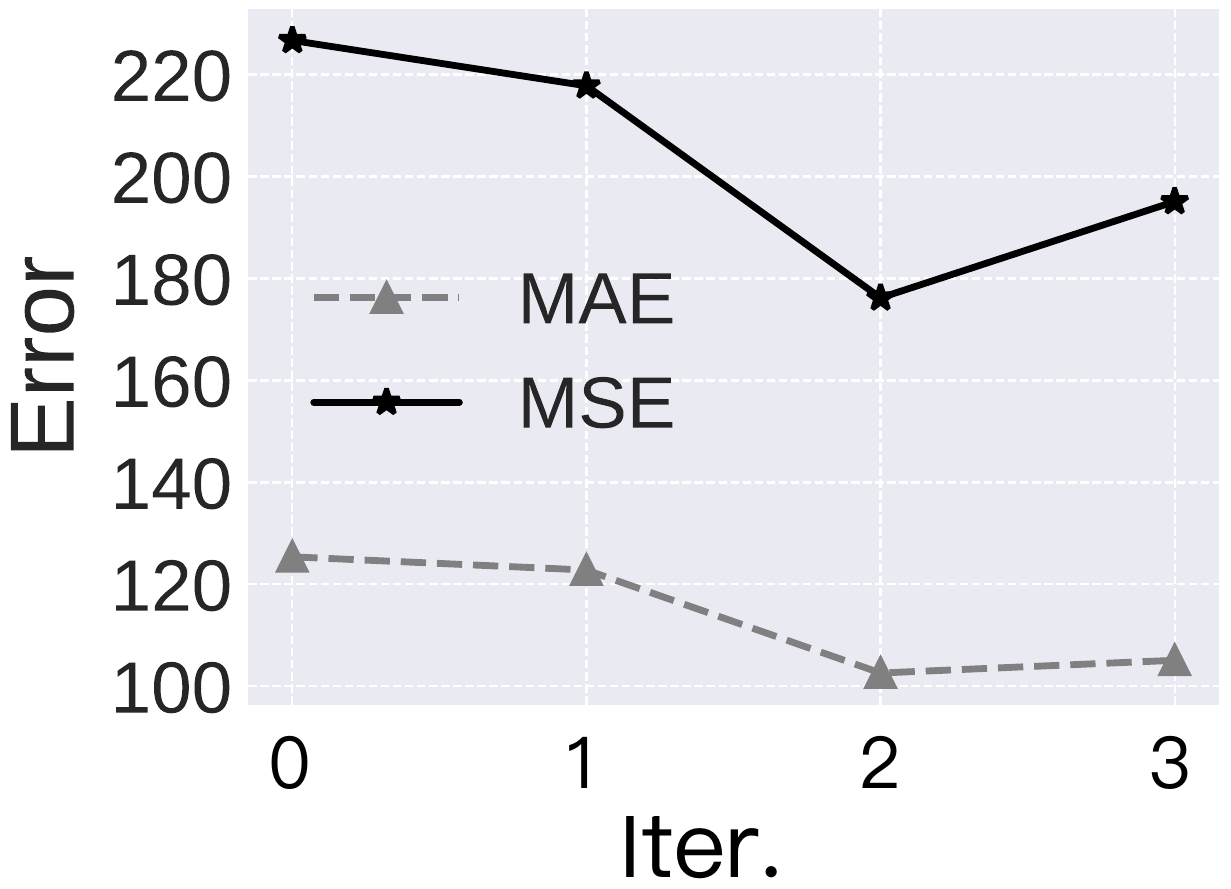} 
        \caption{Error v.s. Iter.}
        \label{fig:stage}
    \end{minipage}
    \begin{minipage}{0.3\textwidth}
        \centering
        \includegraphics[width=\textwidth]{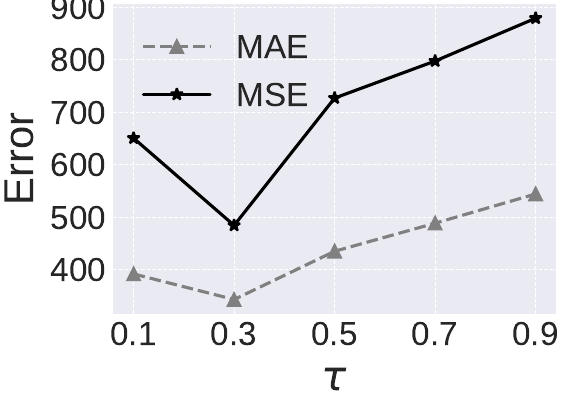} 
        \caption{Error v.s. $\tau$}
        \label{fig:tau}
    \end{minipage}
    \begin{minipage}{0.3\textwidth}
        \centering
        \includegraphics[width=\textwidth]{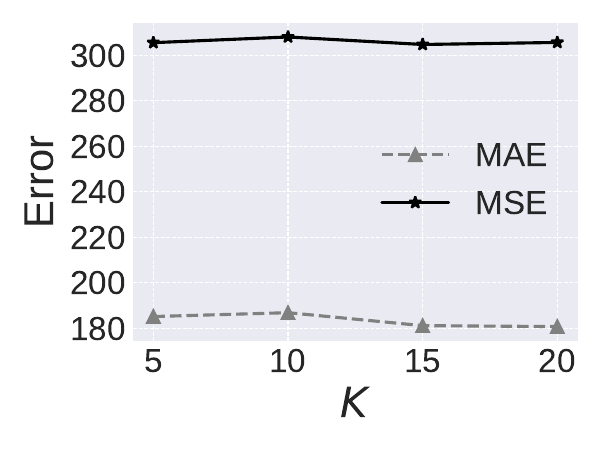} 
        \caption{Error v.s. $K$}
        \label{fig:k}
    \end{minipage}
\end{figure*}

\begin{figure}[t]
    \centering
    \includegraphics[width=0.4\textwidth]{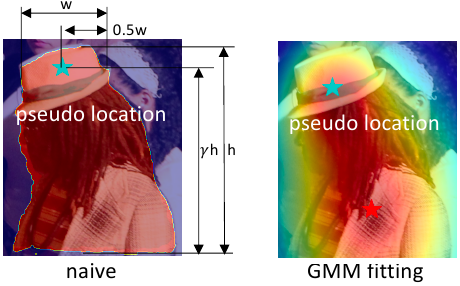}
    \caption{The comparison of naive localization and the proposed robust localization method using GMMs.}
    \label{fig:gmm}
\end{figure}

\textbf{Robust localization}
As shown in Figure~\ref{fig:gmm}, one straightforward approach to localizing head positions is to assume that the ratio of head height to the total height of the mask remains constant. To validate the effectiveness of our proposed GMM fitting method, we compared it against this naive method using various ratios. The results of this comparison are showcased in Figure~\ref{fig:gamma}.

The experiment demonstrates that the GMM fitting method consistently outperforms the naive approach across different ratios. The superiority of the GMM fitting method can be attributed to its ability to learn the dynamic shape of person masks in a data-driven manner. Unlike the naive method, which relies on a fixed and arbitrary assumption about head height ratios, the GMM fitting method adapts to the varying shapes and sizes of individuals in the crowd. This flexibility allows for more accurate and reliable localization of head positions, particularly in diverse and unpredictable crowd scenarios. 

\textbf{Iterative pseudo-label generation}
To enhance mask quality, we introduce an iterative pseudo-labels generation approach, leveraging the trained counting network. First, we predict individual locations in training images, considering local maxima above a threshold as potential person localizations, following \cite{wan2021generalized}. These predicted locations are then used as prompts for SEEM segmentation, effectively localizing new people in dense areas. As Figure~\ref{fig:masks} illustrates, this method detects more people, evidenced by the increased mask count. Performance comparisons in Table~\ref{tab:ablation} show marked improvements with iterative pseudo-label generation by comparing ``AdaSEEM'' and ``AdaSEEM + Iter. 0''. Moreover, ``Ours (Iter. 1)'', a newly trained counting network with these refined masks outperforms the prior iteration, confirming the method's efficacy.

To determine the optimal number of iterations, we experimented on ShanghaiTech A, with results depicted in Figure~\ref{fig:stage}. The findings indicate that peak performance is attained at the second iteration, after which the performance converges.  Consequently, we opted for two iterations in subsequent experiments.

\textbf{Loss hyperparameters}
Figures \ref{fig:omega} and \ref{fig:beta} show the ablation studies for different values of the loss hyperparameters, $\omega$ and $\beta$.
Figure \ref{fig:tau} and \ref{fig:k} shows the ablation study for $\tau$ and $K$.

\textbf{Localization performance}
We further evaluate the localization performance of the proposed method on UCF-QNRF. The performance of our proposed unsupervised method was benchmarked against existing supervised methods, filling a gap as there were no comparable unsupervised crowd localization methods. Despite the lack of manual labeling during training, our method demonstrated commendable precision, which outperforms several supervised counterparts, as shown in Table~\ref{tab:loc}. 
The recall of our method is lower than supervised approaches but can be improved significantly using the 1st iteration training, which confirms that more missed people are detected and pseudo-labeled.
While the overall localization performance of the proposed method is still limited and falls short of the state-of-the-art supervised methods, the results are promising, particularly considering the absence of manual labels. 

\begin{table}[h]
\centering
\resizebox{0.6\textwidth}{!}{
\begin{tabular}{lcccc}
\hline
Method & Label & Precision $\uparrow$ & Recall $\uparrow$ & AUC $\uparrow$ \\
\hline
MCNN \cite{zhang2016single} & Point & 0.599 & 0.635 & 0.591 \\
ResNet \cite{he2016deep}& Point  & 0.616 & 0.669 & 0.612 \\
DenseNet \cite{huang2017densely} & Point & 0.702 & 0.581 & 0.637 \\
Encoder-Decoder \cite{badrinarayanan2017segnet} & Point & 0.718 & 0.630 & 0.670 \\
CL \cite{idrees2018composition} & Point & 0.758 & 0.598 & 0.714 \\
GL \cite{wan2021generalized} & Point & 0.782 & 0.748 & 0.763 \\
\hline
Ours (Iter. 0) & None &0.777&0.101&0.456 \\ 
Ours (Iter. 1) & None &0.677&0.263&0.476 \\ 
\hline
\end{tabular}
}
\caption{Localization performance on UCF-QNRF dataset.}
\label{tab:loc}
\end{table}

\section{Limitation}
\label{sec:limitation}
The current limitation of our proposed method lies in the time-intensive iterative process for pseudo-labels generation, as it requires segmenting all predicted point locations, with the duration increasing with the dataset's population density. To maximize recall, we predict numerous locations, subsequently consolidating overlapping masks using Non-Maximum Suppression (NMS) which further increases the computation time. Future work will focus on developing a more efficient pseudo-label generation technique to enhance training efficiency.

\section{Conclusion}
\label{sec:conclusion}
In our study, we introduce a robust unsupervised crowd counting method that excels in performance compared to previous unsupervised approahces, and rivals some supervised methods. Our approach includes an adaptive resolution SEEM for generating better segmentation masks as pseudo-labels in dense areas, a robust localization technique using GMM fitting on soft masks generated from multiple mask samples, and a counting network trained with a novel loss function excluding uncertain regions. Additionally, we propose an iterative method  to enhance the pseudo labels by using predictions from the well-trained counter to find individuals who have not been pseudo-labeled yet. 
Future work will aim to boost training efficiency and improve localization performance.

\section*{Acknowledgments}
This work was supported by a Strategic Research Grant from City University of Hong Kong (Project No. 7005665).

%
%
\bibliographystyle{splncs04}
\bibliography{main}
\end{document}


\title{Supplemental for Robust Zero-Shot Crowd Counting and Localization With Adaptive Resolution SAM} 

\maketitle

\section{Gaussian Mixture Model for Pseudo Point Labels}
\label{sec:rationale}
A soft mask $M \in \mathbb{R}^{h\times w}$ generated via SEEM can also be represent as $M = \{(s_i, ~x_i)\}_{i=1}^{h\times w}$, where $s_i$ is the score value locating at $x_i$. To find the pseudo point label indicating the human head, we use a mixture of two Gaussian distributions to fit the mask $M$:
\begin{align}
	G = p(x) = \sum_{j=1}^2 \pi_j \mathcal{N}(x|\mu_j, \Sigma_j), \label{eq:gmm}
\end{align}
in which these parameters are estimated effectively through the Expectation Maximization (EM) algorithm in practice. 

In the \emph{E-step}, the soft assignments are computed according to the current estimated $G$. In particular, the likelihood that assigns the
$i$-th score $(s_i, ~x_i)$ to the $j$-th Gaussian distribution is formulated as:
\begin{align}
	\hat{z}_{ij} = p(z_i = j | x_i, G) = \frac{\pi_j\mathcal{N}(x_i|\mu_j, \Sigma_j)}{\sum_{k=1}^2\pi_k\mathcal{N}(x_i | \mu_j, \Sigma_k)}.
\end{align}
After all $\hat{z}_{ij}$ is obtained, the parameters in the two-Gaussian mixture $G$ is updated in \emph{M-step} by maximizing the likelihood:
\begin{align}
	\hat{N}_j &= \sum_{i=1}^{h\times w} s_i \hat{z}_{ij}, \\
	\hat{\pi}_j &= \frac{\hat{N}_j}{\sum_{i=1}^{h\times w} s_i}, \\
	\hat{\mu}_j &= \frac 1 {\hat{N}_j} \sum_{i=1}^{h\times w} s_i \hat{z}_{ij}x_i, \\
	\hat{\Sigma}_j &= \frac{1}{\hat{N}_j} \sum_{i=1}^{h\times w} s_i\hat{z}_{ij}(x_i - \hat{\mu}_j)(x_i - \hat{\mu}_j)^{\top}.
\end{align}

With the estimated parameters, we denote the mean $\hat{\mu}_j$ of the Gaussian component with the smaller vertical coordinate (height) as the head location.

\section{Performance on NWPU-Crowd dataset}

We compare our performance with existing supervised methods on the NWPU-Crowd test set for reference. The result is shown in Table~\ref{tab:nwpu}. The proposed method achieves comparable performance to some supervised models.

\begin{table}[!h]
\footnotesize
\vspace{-7mm}
  \caption{Comparison with supervised methods on NWPU (test).}
  \label{tab:nwpu}
  \centering
  \resizebox{0.48\textwidth}{!}{%
  \begin{tabular}{l cccccc}
    \toprule
    {Method} &Label& MAE $\downarrow$ & MSE $\downarrow$ & Prec & Rec & F1 \\
    \cmidrule(r){1-7}
    TinyFaces &Point&272.4 &764.9&0.529&0.611&	
0.567 \\
    MCNN &Point&232.5&714.6 &-&-&- \\
    SANet &Point&190.6&491.4 &-&-&- \\
    GeneralizedLoss &Point&79.3&346.1&0.800&0.562&0.660 \\
    Ours  &None&168.4&547.5&0.762&0.510&0.611 \\
    \bottomrule
  \end{tabular}
}
\vspace{-4mm}
\end{table}